\documentclass[10pt,letterpaper]{article}
\usepackage[top=0.85in,left=2.75in,footskip=0.75in]{geometry}

\usepackage{amsmath,amssymb}

\usepackage{changepage}

\usepackage[utf8x]{inputenc}

\usepackage{textcomp,marvosym}

\usepackage{cite}

\usepackage{nameref,hyperref}

\usepackage{microtype}
\DisableLigatures[f]{encoding = *, family = * }

\usepackage[table]{xcolor}

\usepackage{array}

\newcolumntype{+}{!{\vrule width 2pt}}

\newlength\savedwidth

\raggedright
\usepackage[aboveskip=1pt,labelfont=bf,labelsep=period,justification=raggedright,singlelinecheck=off]{caption}

\makeatletter
\renewcommand{\@biblabel}[1]{\quad#1.}
\makeatother

\usepackage{lastpage,fancyhdr,graphicx}
\usepackage{epstopdf}
\fancyheadoffset[L]{2.25in}
\fancyfootoffset[L]{2.25in}
\begin{document}
\vspace*{0.2in}

% Title must be 250 characters or less.
\begin{flushleft}
{\Large
\textbf\newline{A Deep Learning Generative Model Approach  \\ [2mm] for Image Synthesis of Plant Leaves} % Please use "sentence case" for title and headings (capitalize only the first word in a title (or heading), the first word in a subtitle (or subheading), and any proper nouns).
}
\newline
% Insert author names, affiliations and corresponding author email (do not include titles, positions, or degrees).
\\
Alessandro Benfenati \textsuperscript{1},
Davide Bolzi\textsuperscript{2},
Paola Causin \textsuperscript{2*},
Roberto Oberti \textsuperscript{3}
\\
\bigskip
\textbf{1} Dept. of Environmental Science and Policy, Universit\`a degli Studi di Milano, Milano, Italy
\\
\textbf{2} Dept. of Mathematics, Universit\`a degli Studi di Milano, Milano, Italy
\\
\textbf{3} Dept. of Agricultural and Environmental Sciences - Production, Landscape, Agroenergy, Universit\`a degli Studi di Milano, Milano, Italy
\\
\bigskip

% Insert additional author notes using the symbols described below. Insert symbol callouts after author names as necessary.
% 
% Remove or comment out the author notes below if they aren't used.
%
% Primary Equal Contribution Note
%\Yinyang Author contributions: A.B and P.C. designed research; A.B., D.B and P.C. performed research; R.O. provided the plants, %provided expertise and analyzed the results; A.B. and P.C. wrote the paper

% Additional Equal Contribution Note
% Also use this double-dagger symbol for special authorship notes, such as senior authorship.
%\ddag These authors also contributed equally to this work.

% Current address notes
%\textcurrency Current Address: Dept/Program/Center, Institution Name, City, State, Country % change symbol to "\textcurrency a" if more than one current address note
% \textcurrency b Insert second current address 
% \textcurrency c Insert third current address

% Deceased author note
%\dag Deceased

% Group/Consortium Author Note
%\textpilcrow Membership list can be found in the Acknowledgments section.

% Use the asterisk to denote corresponding authorship and provide email address in note below.
* paola.causin@unimi.it

\end{flushleft}
% Please keep the abstract below 300 words
\section*{Abstract}

\subsection*{Objectives}
We generate via advanced Deep Learning (DL) techniques artificial 
leaf images in an automatized way.
Our aim is to dispose of a source of training samples in artificial intelligence applications
for modern crop management in agriculture, with focus on disease recognition
on plant leaves.
Such applications require large amounts of data 
and, while leaf images are not truly scarce, image collection and annotation 
remains a very time--consuming process.
Data scarcity can be addressed
by augmentation techniques consisting in 
simple transformations of samples belonging to a small dataset, but the richness 
of the augmented data is limited: this motivates the search for alternative approaches.

\subsection*{Methods}
Pursuing an approach based on 
DL generative models,
we propose a Leaf-to-Leaf
Translation (L2L) procedure structured in two steps: firstly, a residual variational 
autoencoder architecture is used to generate synthetic leaf skeletons (leaf profile and veins)
starting from companions binarized skeletons of real leaf images. In a second step, we perform the process of translation via a {\tt Pix2pix} framework,
which uses conditional generator adversarial networks
to reproduce the colorization of leaf blades, preserving the
shape and the venation pattern.  

\subsection*{Results} 
The L2L procedure generates synthetic images of leaves with 
a realistic appearance. We address the performance measurement both in a qualitative and a quantitative way; for this latter evaluation, we employ a DL anomaly detection strategy which 
quantifies the degree of anomaly of synthetic leaves with respect to real samples. 

\subsection*{Conclusions} 
Generative DL approaches have the potential to be a new paradigm 
to provide low-cost meaningful synthetic samples 
for computer-aided applications. 
The present L2L approach represents a step towards this goal, being able to generate synthetic
samples with a relevant qualitative and quantitative resemblance to real leaves.

% Please keep the Author Summary between 150 and 200 words
% Use first person. PLOS ONE authors please skip this step. 
% Author Summary not valid for PLOS ONE submissions.   
\section*{Author summary}
In this work we present an end-to-end workflow incorporating state-of-the-art 
Deep Learning strategies based on generative methods to produce realistic synthetic images of leaves. At the best of our knowledge, this is the first attempt of such an 
approach to this problem. Our focus application is the training of 
neural networks for modern crop management systems in agriculture,
but we believe that many other computer--aided applications may benefit from it.
We take inspiration from previous works carried out on 
eye retina image synthesis, an application domain which shares some similarities with the present problem
(a venation pattern over a colorized ``fundus''). Our approach relies on the  
successive use of autoencoders and generative adversarial architectures, able to generate leaf
images both in the Red-Green-Blue channels as well as in the Near-Infra-Red.
The generated leaf images have a realistic appearance even if they sometimes suffer from small inconsistencies, especially discolored patches. A quantitative evaluation via an anomaly detection
algorithm shows that in average a synthetic sample is classified as such only in 25\% of the cases.
%\linenumbers

% Use "Eq" instead of "Equation" for equation citations.
\section*{Introduction}
The ability to generate meaningful synthetic images of leaves 
is highly desirable for many computer-aided applications.
At the best of our knowledge, attempts at generating synthetic images of leaves
have been made mostly in the field of computer graphics 
and were aimed at creating the illusion of realistic landscapes covered with plants, trees or meadows.
These efforts were mainly based on 
procedural mathematical models describing the venation structure and the color/texture of the leaf.
A specific type of formal grammar, called L--grammar, was developed 
to generate instructions to draw a leaf. 
Several instances of the profile of leaves of a certain species were created upon random variations of parameters of the directives of a certain L--grammar~\cite{peyrat2008generating}.
Biologically-motivated models were also proposed. A main point of these approaches
is the representation of the interaction between auxin
sources distributed over the leaf blade and the formation of 
vein patterns~\cite{runions2005modeling}.  
Some attempts were also carried out using finite elements to build mechanistic models
of the leaf blade, tuned on its structural parameters~\cite{samee2012modeling}.
After generating the leaf shape and venation pattern - regardless of the approach-  texture and
colors were rendered by a color palette prescribed by the user or generated according to a pseudo--random algorithm.
A color model based on convolution sums of divisor functions 
was proposed in~\cite{kim2017procedural}, while 
a shading model based on the PROSPECT model for light transmission in 
leaves~\cite{feret2017prospect}, was 
proposed in~\cite{miao2013framework}. ``Virtual rice''  leaves 
were created in~\cite{yi2016modeling} based on a RGB-SPAD model.
%For example, the authors of~\cite{kim2017procedural} proposed a color model based on convolution sums of divisor functions, the authors of~\cite{miao2013framework} used a biologically--based shading model for rendering of plant leaves based on the different contents of chlorophyll and carotenoids, according to the PROSPECT model for light  transmission~\cite{feret2017prospect} and the authors of~\cite{yi2016modeling} created ``virtual rice''  leaves  based on RGB-SPAD model.

In this work we aim at introducing a radically different approach by
generating artificial images of leaves by automatized techniques based on 
Deep Learning (DL) techniques.  
Our focus is mainly to enrich dataset of leaf images for
neural networks training,
even if we deem that the present approach may be of interest also in a wide range of other 
fields, starting from computer graphics.
%In the application we have in mind, DL techniques allow to implement efficient survey strategies of crops to spot early localized foci of diseases on leaves (see, {\em e.g.},  \cite{ferentinos2018deep, saleem2019plant}). 
The motivation underlying this work is that DL methods require a large amount of data --  often of the order of hundreds of thousands of images --  to avoid
overfitting phenomena. Data augmentation is a common remedy, usually
consisting in simple transformations such as random rotations, translations
or deformations of the original images. However, the richness of the augmented dataset is limited and more sophisticated approaches for
synthesizing additional training data have a greater potential to improve the training process. 
In this respect, DL generative models 
represent attractive methods to produce synthetic images
(with corresponding labels) using the information
from a limited set of real, unlabeled images of the same domain.
This idea is not new - in absolute - but it has been used mainly
in the field of medicine, where data may be extremely
scarce and difficult to obtain (see, {\em e.g.}, the recent review~\cite{taghanaki2021deep}).
In the present context, while scarcity of leaf images may be not a real issue, what is more relevant is 
to avoid the huge mole of work required to collect,
examine and annotate images. 
This is especially true when image segmentation should be
performed, which is a pixel-wise problem: the acquisition of annotated 
segmentation masks is exceedingly costly and time consuming, as a human expert annotator has to label
every pixel manually.
For our model we take inspiration from \cite{costa2017end} (and reference therein), where the authors synthesised eye retina images.
%For our model we take inspiration from the works by Costa and co-authors~(see~\cite{costa2017end} and references therein) carried out on  the eye retina image synthesis. 
The fundus of the eye
shares indeed several characteristics with our problem:
a fine network of hierarchically
organized blood vessels (leaf veins) superposed to  
a colored background (leaf blade). In addition, in our problem  
the leaf blade is also characterized by a specific silhouette that must be represented as well.  
We propose a Leaf-to-Leaf
Translation (L2L) approach to obtain synthetic colorized leaf blades organized in two steps: first 
we use a residual variational 
autoencoder architecture to generate fake leaf skeletons starting from binarized companion skeletons of real leaf images. 
In a second step we perform the process of translation
via a {\tt Pix2pix} framework,
which uses conditional generator adversarial networks
(cGANs) to reproduce the specific color distribution of the leaf blade, preserving leaf
shape and venation pattern.  
We carry out both qualitative and quantitative evaluations of the degree of realism of the synthetic
samples of leaves. Specifically, a DL-based anomaly detection strategy is used to evaluate
the distance (``anomaly'') between synthetic and real samples. The results show a
good degree of realism, that is a low anomaly score, and indicate that with the present approach
one can significantly enrich a small dataset and improve the training performance
of DL architectures.  

\section*{Materials and methods}

\subsection*{Dataset}
\begin{figure}[hp]
\centering
\includegraphics[width=\linewidth]{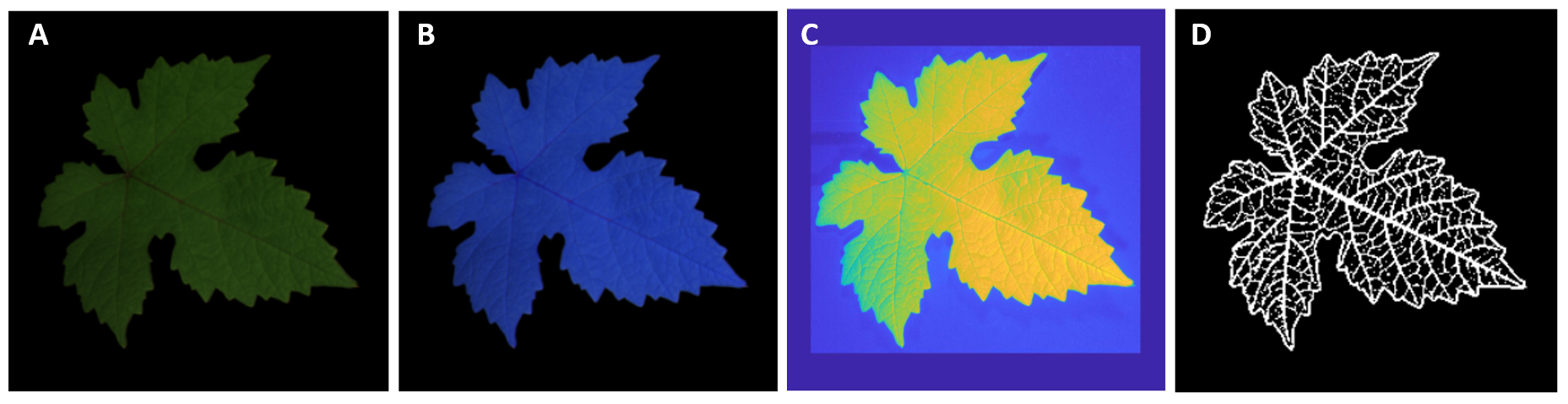}
\caption{{\bf Sample of grapevine leaf from the dataset.} A: RGB image; B: RGNIR image;  
C: normalized and cropped NIR image; D: companion skeleton. In the skeleton
binarized image, the white color identifies the leaf profile and veins, the black color identifies
other parts of the leaf and the background.}
\label{fig:start}
\end{figure}

Grapevine leaves were imaged via 
a QSi640 ws-Multispectral camera (Atik Cameras, UK) equipped with a Kodak 4.2~Mp micro-lens image sensor and 
8 spectral selection filters operating in the bands 430 to 740~nm. 
For the purpose of this experiment, leaves were imaged singularly on a dark background, under controlled diffuse illumination conditions. 
Images were acquired in the single spectral channels 
430~nm (blue, B), 530~nm (green, G), 685~nm (red, R) and 740~nm (near--infrared, NIR). 
These channels are typically considered when dealing with the task of recognition of plant diseases
in a multispectral analysis approach~\cite{oberti2014automatic,mahlein2018hyperspectral}. 
A set of RGB images of the same leaves in standard CIE color space
were also acquired for reference. 
Camera parameters were set and 
image collection was performed via an {\em in--house} 
developed acquisition software written in MATLAB. 
Reflectance calibration was carried out by including in each image 3 reflectance references
 (Spectralon R = 0.02, R = 0.50 and R = 0.99; Labsphere, USA).
 We obtained photos of 80 leaves with a resolution of $2048 \times 2048$ pixels
and 8 bit for each channel. 
Preprocessing operations were performed on each image: removal of {\em hot} pixels, 
normalization along each channel according to the reference probes, 
creation of a companion binarized skeleton image.
For this latter procedure, the NIR channel was used, since it presents an a high contrast
between the leaf and background. 
The skeleton comprises the profile of the leaf and the vein pattern.
Images and companion skeletons were resized at $256 \times 256$ resolution.
Fig~\ref{fig:start} shows the original images in the RGB and RGNIR spaces, the 
normalized NIR channel and the corresponding companion skeleton.
Before using the generative algorithms, we performed 
standard data augmentation by randomly flipping each image horizontally and vertically, rotating by an angle randomly chosen in $[-\pi/4,\pi/4]$ and finally zooming with a random amount in the range $[-20\%,+20\%]$. The dataset was thus increased in this way from 80 to
240 samples.

\subsection*{Generative methods for L2L translation}

The authors of~\cite{costa2017end,sengupta2020funsyn} generated
artificial patterns of blood vessels along with corresponding eye fundus images 
using a common strategy which divides the problem of the image generation into two sub--problems, 
each one addressed by a tailored DL architecture: first they generate the blood vessel tree,
then they color the eye fundus. We adopt this very approach, first generating the leaf profile and veins and
then coloring the leaf blade. Also in our experience this approach
has turned out  to be more effective than generating the synthetic image altogether.  

\subsubsection*{Skeleton Generation}

\begin{figure}[h]
\centering
\includegraphics[width=.8\linewidth]{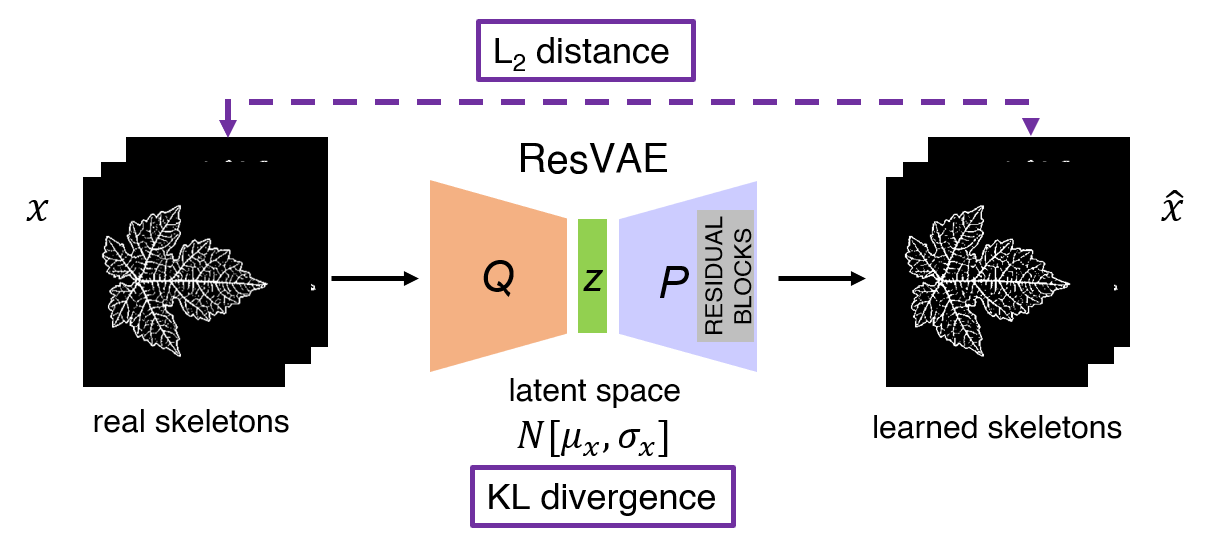}
\caption{{\bf  Illustration of the  ResVAE framework (training)}.}
\label{fig:ResVAEschema}
\end{figure}

According to the above considerations, the generation of a realistic leaf skeleton is the first
step towards the final goal of our work. 
For this task, we use a convolutory autoencoder architecture, 
that is, a network trained to reconstruct its input.
An autoencoder (AE) is composed of two submodels: 1) an encoder $Q$ that
maps the training dataset to a latent (hidden) representation
$z$; 2) a decoder $P$ that maps $z$ to an output that
aims to be a plausible replica of the input. 
We have experimented that simple autoencoders 
cannot generate realistic skeletons.  
For this reason, we use a more sophisticated architecture, called
Residual Variational Autoencoder (ResVAE, see Fig~\ref{fig:ResVAEschema}).  

This learning framework has already been successfully
applied to image recognition, object detection, and image super-resolution (see, {\em e.g.},~\cite{cai2019multi}). 
In the data generation framework, AEs learn the projection of the initial data into a \emph{latent subspace}, and then a sample of this subspace is randomly extracted to build up a new instance of the initial data. Instead of learning such projection, VAEs learn the probability distribution of the latent variables given the input $x$. 
As a matter of fact, a variational autoencoder can be defined as an autoencoder whose training is regularized to avoid overfitting and ensure that the latent space has good properties that enable the generative process.
To achieve this goal, instead of encoding an input as a single point, VAEs encode it as a 
(Gaussian) distribution over the latent space,
where $p(z|x)$ represents the probability of the latent variable $z$ given the input $x$. The decoding part consists in sampling a variable from $p(z|x)$ and then providing a reconstruction $\widehat{x}$ of the initial data $x$. 
We associate to this framework the following loss function
%\begin{equation}
%\mathcal{L}_{\text{VAE}} = 
%\mathcal{L}_{L_2} + \beta \mathcal{L}_{KL},  
%\label{eq:lossVAE}
%\end{equation}
\begin{equation}
\mathcal{L}_{\text{VAE}}(x,\hat{x}) = 
\mathcal{L}_{L_2}(x,\hat{x}) + \beta \mathcal{L}_{KL}\left( p(z|x),\mathcal{N}(0,1)\right),  
\label{eq:lossVAE}
\end{equation}
where the first term $\mathcal{L}_{L_2} =||x-\widehat{x}||^2$  is the $L_2$ norm of the reconstruction loss,  and the second term $\mathcal{L}_{KL}=KL[N(\mu_x,\sigma_x),N(0,1)]$
is the Kullback--Leibler (KL) divergence~\cite{asperti2020balancing,benfenati2015image}
The KL divergence enhances sparsity in neurons activation to improve the quality of the latent features
keeping the corresponding distribution close to the Gaussian distribution $\mathcal{N}(0,1)$. The tunable regularization hyperparameter $\beta$ is 
used to weigh the two contributions\cite{higgins2016beta}.  With respect to VAEs, ResVAEs additionally employ residual blocks and connection skips. The idea beyond residual blocks is the following~\cite{he2016deep}: normal layers try to directly learn an underlying mapping, say $h(x)$, while residual ones approximate a residual function $r(x) = h(x)- x$. Once the learning is complete, $r(x)$ is added to the input to retrieve the mapping: $h(x) = r(x)+x$. 
In our architecture, residual blocks are concatenated to the decoder to increase the capacity of model~\cite{cai2019multi}.
The connection skips allow to back--propagate the gradients more efficiently
giving the bottleneck more access to the simpler features extracted earlier in the encoder.
 The resulting ResVAE compresses $256\times 256$ leaf skeleton images
to a low dimension latent vector of size 32 and then it reconstructs it 
to $256\times 256$ images.
We refer to~\nameref{SI:ResVAE} for specifications of the present ResVAE architecture.

%\begin{table}%[tbhp]
%\centering
%\caption{Specification of the ResVAE architecture. Abbreviations: Conv\,=\,convolution; TConv\,=\,transposed convolution; BN\,=\,BatchNormalization;
%LR($x$)\,=\,LeakyReLU activation with parameter $x$; $kz_1nz _2$: kernel with size\,=\,($z_1 \times  z_1$), feature map with $z_2$ features.}
%\begin{tabular}{lrr}
%Encoder & \\
%Layer type & kernel type  \\
%\midrule
% Network input (256 $\times$ 256 $\times$ 1) &  - \\
% BN+Conv+BN+LR(0.2) & k2n16 \\
% Conv + BN+ LR(0.2) & k1n16 \\
% Conv + BN+ LR(0.2) & k2n32 \\
% Conv + BN+ LR(0.2) & k2n64 \\
% Conv + BN+ LR(0.2) & k1n64 \\
% Conv + BN+ LR(0.2) & k2n128 \\
% Conv + BN+ LR(0.2) & k1n128 \\
% Conv + BN+ LR(0.2) & k2n256 \\
% Conv + BN+ LR(0.2) & k1n256 \\
% Flatten & - \\
% Dense32 ($\mu$), Dense32 ($\sigma$) & \\ 
% Sampling & - \\
%\bottomrule
%\end{tabular}
%\qquad 
%\begin{tabular}{lrr}
% Decoder & \\
% Layer type & kernel type & skip connection \\
%\midrule
% TConv + BN+ LR(0.2) & k2n256 &  \\
% Conv + BN+ LR(0.2) & k1n256 & $\times$ \\
% TConv + BN+ LR(0.2) & k2n128 \\
% Conv + BN+ LR(0.2) & k1n128 & $\times$ \\
% TConv + BN+ LR(0.2) & k2n64 \\
% Conv + BN+ LR(0.2) & k1n64 & $\times$ \\
% TConv + BN+ LR(0.2) & k2n32 \\
% Conv + BN+ LR(0.2) & k1n32 & $\times$ \\
% TConv + BN+ LR(0.2) & k2n16 \\
% Conv + BN+ LR(0.2) & k1n16  & $\times$\\
%Residual block & - & \\
% Conv + Sigmoid & k2n1 & \\
%\bottomrule
%\end{tabular}
%\label{tab:ResVAE}
%\end{table}

\subsubsection*{Translation to colorized leaf image}

\begin{figure}[h]
\centering
\includegraphics[width=1.05\linewidth]{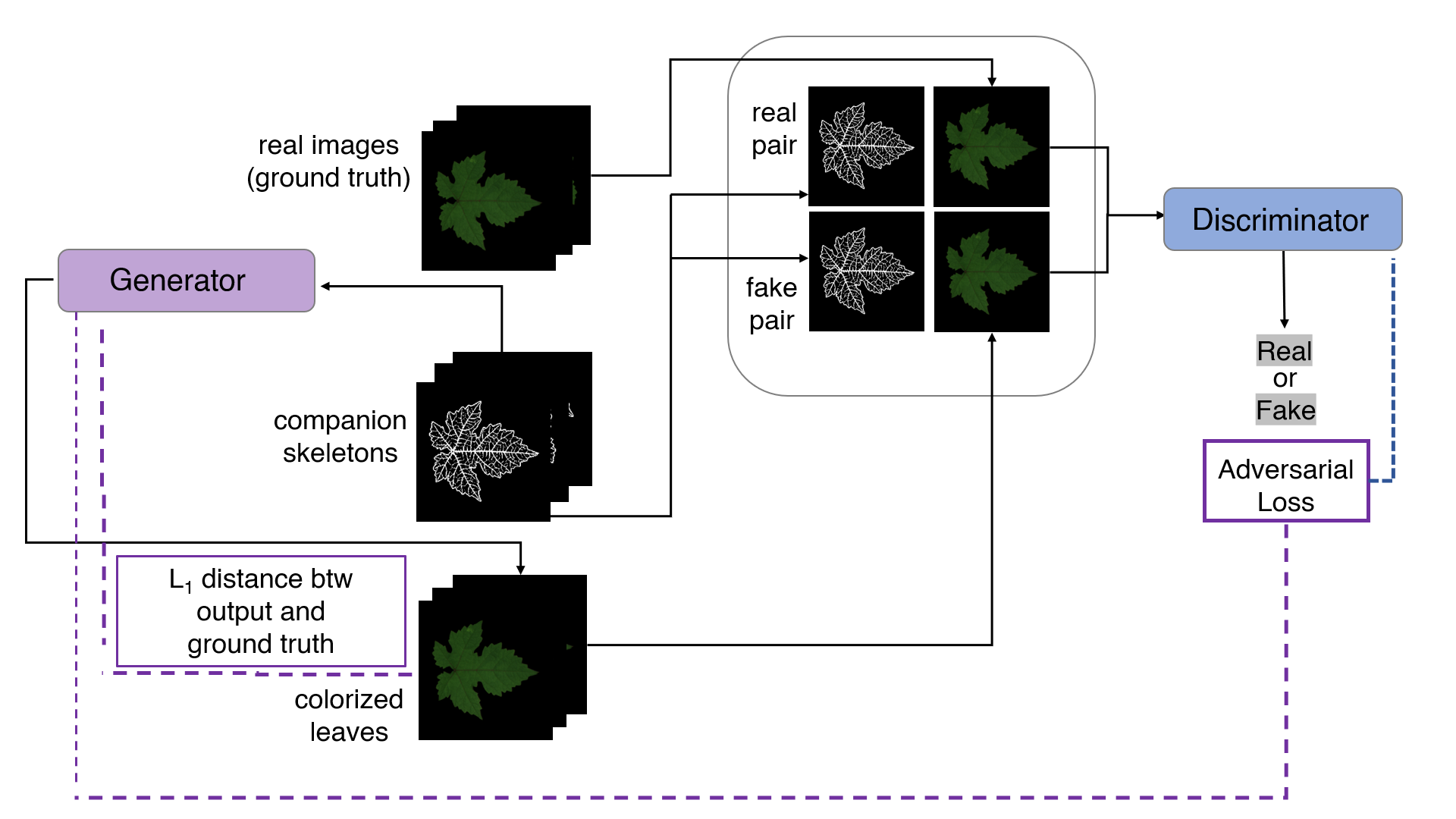}
\caption{{
\bf Illustration of the {\tt Pix2Pix} framework (training).}}
\label{fig:schemaPix2pix}
\end{figure}

We consider the colorization of the leaf out of an existing 
skeleton as an image-to-
image translation problem, which implies to learn a mapping from
the binary vessel map into another representation.
Similarly to what observed in~\cite{costa2017end} for retinal image generation, 
many leaf images can share a 
similar binary skeleton
network due to variations in color, texture, illumination. For this
reason, learning the mapping   
is an ill-posed problem and some uncertainty
is present.
%The strategy we pursue to learn the mapping adopts a {\tt Pix2pix} net (also known cGAN), an unsupervised generative model which includes two deep neural networks, a generator $G$ and discriminator $D$.
We learn the mapping via a  {\tt Pix2pix} net (also known as conditional GAN, (cGAN)), an unsupervised generative model which represents a variation of a standard GAN.
As such it
includes two deep neural networks, a generator $G$ and discriminator $D$.
The generator aims to capture the data distribution, while
the discriminator estimates the probability that a sample actually
came from the training data rather than from the generator.
In order to learn a generative distribution over the data $x$, 
the generator builds a mapping $G(z; \theta_G)$ from a prior noise distribution $p_z$
to the image data space, $\theta_G$ being the
generator parameters. The discriminator outputs 
the probability that $x$ came from the real data distribution $p_{data}(x)$ rather from 
the generated one.
We denote by $D(x; \theta_D)$ the discriminator function,
 $\theta_D$ being the
discriminator parameters. In standard GANs, the optimal mappings $G^*$ is obtained as the equilibrium point of the min--max game:
$$
(G^*,D^*) =\arg \displaystyle\min_{G} \max_D  \mathcal{L}_{GAN}(D,G),
$$
where we have defined the objective function
\begin{equation}
\mathcal{L}_{GAN}(D,G):=
\mathbb{E}_{x \sim p_{data}(x)}[\log D(x; \theta_D)]+
\mathbb{E}_{z \sim p_{z}(z)}[\log (1-D(G(z; \theta_G)))].
\label{eq:lossGAN}
\end{equation}
In the conditional framework, an extra variable $y$ is added as a further source of information on $G$, which combines the noise prior $p_z(z)$ and~$y$.
% In the conditional framework, an extra information~$y$ is provided to condition (generally)~$G$, which combines the noise prior $p_z(z)$ and~$y$. 
The objective function thus becomes
\begin{equation}
\mathcal{L}_{cGAN}(D,G)=
\mathbb{E}_{x \sim p_{data}(x)}[\log D(x; \theta_D)]+
\mathbb{E}_{z \sim p_{z}(z)}[\log (1-D(G(z|y; \theta_G)))].
\label{eq:losscGAN}
\end{equation}
Previous approaches have found it beneficial to mix the
GAN objective with a more traditional loss, such as $L_2$ distance~\cite{kurach2019large}. 
The discriminator’s job remains unchanged, but
the generator is bound not only to fool the discriminator but
also to stay near the ground truth output in an $L_2$ sense. In this work we
rather explore the use of the $L_1$ distance rather than $L_2$ as
$L_1$ promotes sparsity and at the same time it encourages less blurring~\cite{isola2017image}:
\begin{equation}
\mathcal{L}_{L_1}(G)=
\mathbb{E}_{x,y,z}[||y-G(z|y; \theta_G)||].
\label{eq:L1}
\end{equation}
The final objective is thus
\begin{equation}
(G^*,D^*)=\arg \min_{G} \max_D  \mathcal{L}_{cGAN}(D,G)+
\lambda \mathcal{L}_{L_1}(G)
\label{eq:losscGANlambda}
\end{equation}
 where $\lambda$ is a regularization hyperparameter.
In our implementation the extra information corresponds to the leaf skeletons 
which condition~$G$ in the image generation task to preserve leaf shape and
venation pattern.
The discriminator
is provided with skeleton plus generated image pairs and must determine
whether the generated image is a plausible (feature preserving) translation.
Fig~\ref{fig:schemaPix2pix} shows the training process of the cGAN.
We refer to~\nameref{SI:Pix2pix} for specifications of the {\tt Pix2pix} architecture we adopted. 

\subsubsection*{L2L workflow: from random 
samples to leaf images}
\begin{figure}[h]
\centering
\includegraphics[width=\linewidth]{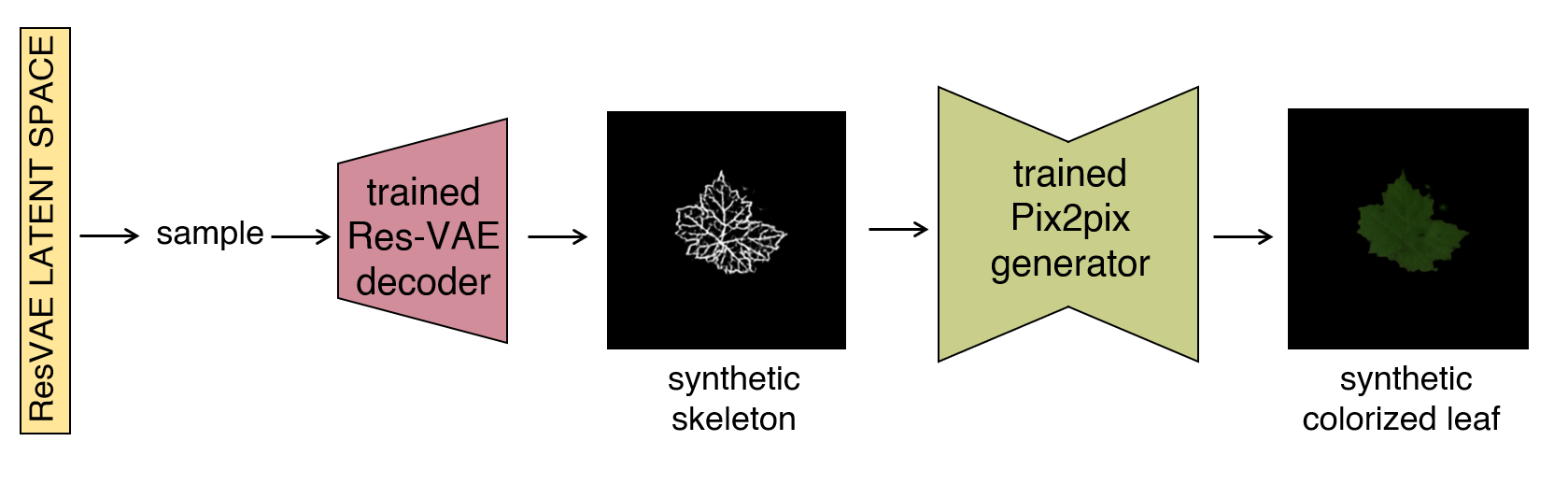}
\caption{{
\bf L2L workflow illustration}. A random input vector is drawn from the ResVAE latent space representation and is input into the trained ResVAE decoder. This latter outputs a synthetic leaf skeleton, which in turn is fed into the trained generator of the {\tt Pix2Pix} and translated into a corresponding colorized leaf.}
\label{fig:L2Lmodel}
\end{figure}

Upon training of the ResVAE and {\tt Pix2pix} architectures, we dispose of an end-to-end procedure
for the generation of synthetic leaves. The procedure, which is completely unsupervised,  
can be summarized as follows (see also Fig~\ref{fig:L2Lmodel}): 
\begin{enumerate}
\item Load weights of the trained ResVAE decoder and {\tt Pix2pix} generator.
\item Draw a random vector from a normal distribution whose parameters are chosen
according to the ResVAE latent space representation (note that its size equals the dimension of the latent space used in the ResVAE, 32 in the present case). 
 \item Input the random vector in the trained ResVAE decoder and generate a leaf skeleton
 \item Input the leaf skeleton into the trained 
 generator of the {\tt Pix2Pix} net to translate it into a fully colorized leaf. 
\end{enumerate}

\section*{Results and Discussion}

The proposed technique can be employed to generate as
many synthetic leaf images as the user requires. 
The model has been implemented with Keras~\footnote{Code to reproduce our experiments will be made available upon publication of this work.}. 
Upon generation of the synthetic images, their quality is assessed  
performing both experimental qualitative
(visual) and quantitative evaluations as follows.  

\subsection*{Visual qualitative evaluation}
\begin{figure}[ht]
\centering
\includegraphics[width=1.05\linewidth]{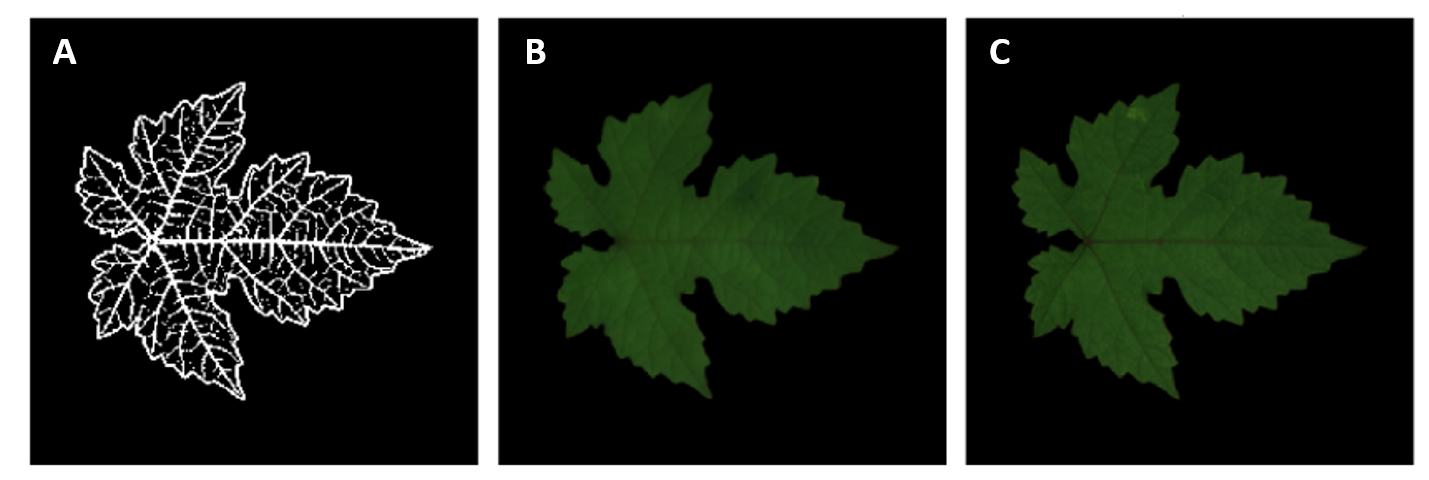}
\caption{{
\bf Consistency test}. The companion binarized leaf skeleton of a real leaf is passed through the generator of the {\tt Pix2Pix}
net to check whether the synthetic colorized leaf blade is similar to the original one. A: companion skeleton of a real leaf; B: 
synthetic colorized blade generated; C: original leaf.}
\label{fig:consistency}
\end{figure}

\noindent{\em Consistency test.} Beforehand, we have evaluated the consistency of the methodology by
verifying that the net has learned to translate a leaf sample comprised in the training set into itself. 
Fig~\ref{fig:consistency} shows an example of this test. The generated leaf is very similar to the real one,  except for some vein discoloration and a small blurring effect effect,  which is a well--known product of AEs employed in image generation~\cite{huang2018introvae}.

\medskip

\begin{figure}[h]
\centering
\includegraphics[width=0.68\linewidth]{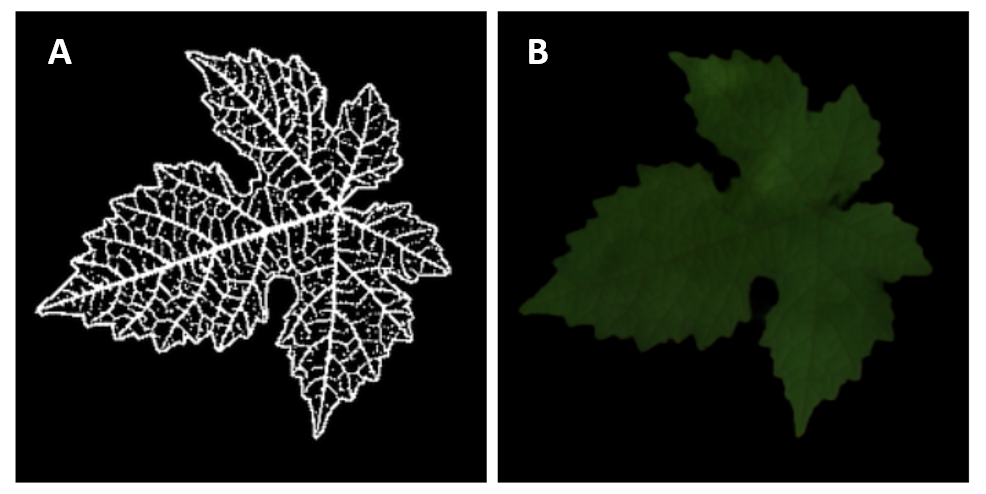}
\caption{{
\bf Translation from unseen real companion skeleton.} A binarized leaf skeleton companion of a real leaf not belonging to
the training set is passed through the generator of the {\tt Pix2Pix}
net to check. A: companion skeleton; B: 
synthetic colorized blade.}
\label{fig:real_skeleton}
\end{figure}

\noindent{\em Translation from unseen real companion skeleton.} Having ensured that the model has learned 
to translate on the training data, we verify that it is able to produce reliable synthetic images using skeletons obtained from leaves that are not part of the training dataset.
%we verify that it is able to make good predictions leaf skeletons obtained  from real leaves not belonging to the training set. 
Fig~\ref{fig:real_skeleton} shows an instance of colorized leaf
obtained from this test. 

\medskip

\begin{figure}[ht] 
\centering
\includegraphics[width=0.68\linewidth]{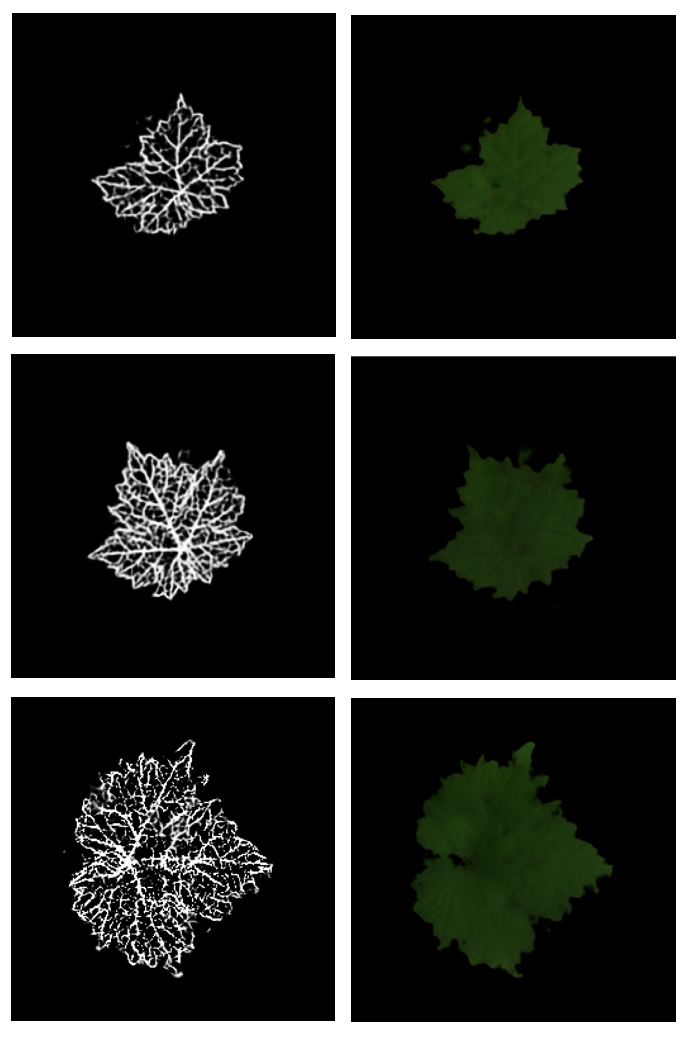}
\caption{{
\bf Full L2L translation results.} Examples of synthetic colorized leaves along with the corresponding synthetic companion skeletons.}
\label{fig:real_leaves}
\end{figure}

\noindent{\em Full L2L translation}  Fig~\ref{fig:real_leaves} shows several instances of synthetic colorized leaves obtained
starting from different random latent vectors. Note that
the generated leaf images differ in terms of their global appearance,
that is the model generalizes and does not trivially memorizes the examples.
As a note, one should observe that some discolored parts may be appear. Moreover, sometimes the skeletons show 
small artifacts consisting in not--connected 
pixels positioned outside the leaf boundary  (not appearing in Fig~\ref{fig:real_leaves}). This latter issue will be addressed via a refinement algorithm explained below.  

\medskip

\begin{figure}[h]
\centering
\includegraphics[width=0.65\linewidth]{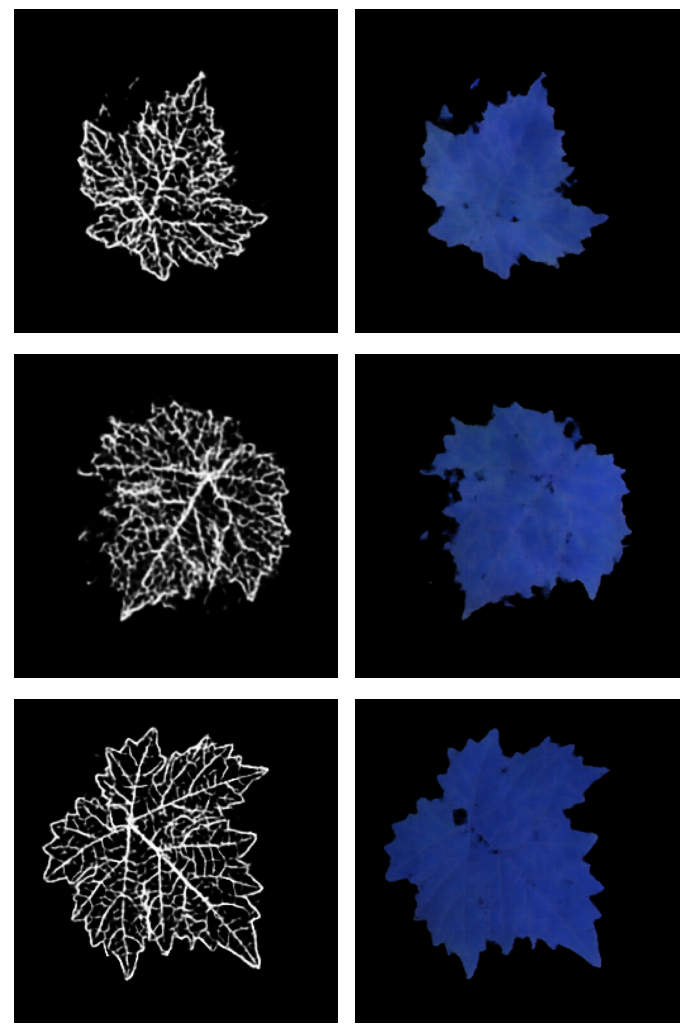}
\caption{{
\bf L2L-RGNIR translation results.} Examples of synthetic leaves colorized in the RGNIR channels along with the corresponding synthetic companion skeletons.}
\label{fig:real_NIR}
\end{figure}

\noindent{\em L2L-RGNIR translation}
As mentioned above, applications in crop management require to have at disposal images 
also in the NIR channel. To do this, we use the L2L generation procedure as for the RGB channels 
starting from RGNIR 
images as {\tt Pix2Pix} targets. Since the same leaf skeletons are used, it 
is not necessary to re-train the ResVAE if this procedure has been already carried out for the RGB
case.  
Fig~\ref{fig:real_NIR} shows some results of this model. 

\medskip

\begin{figure}[h]
\centering
\includegraphics[width=0.8\linewidth]{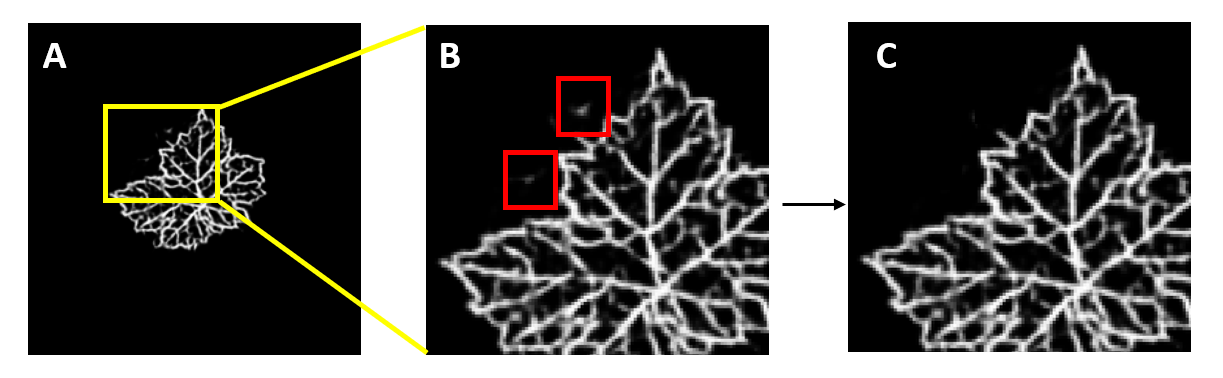}
\caption{{
\bf Refinement algorithm.} The generative procedure sometimes produces artifacts, that is leaf regions
that appear outside the leaf blade. These artifacts are corrected by procedurally finding the contours of all 
the objects in the image and removing the objects outside the leaf contour. A: first leaf in Fig~\ref{fig:real_leaves}
presenting artifacts; B: inset showing the magnified artifacts; C: cleaned leaf.}
\label{fig:refin}
\end{figure}

\noindent {\em Refinement algorithm.} 
We have already discussed the fact that synthetically generated images may 
sometimes present artifacts (leaf regions that appear detached from the leaf blade). 
Obviously this is not realistic and we need to remove such artifacts. 
The refinement algorithm is implemented at present in a procedural way and it is based
on the idea of finding the contours of all the objects and removing all objects laying outside the leaf contour. 
%Observe that in this procedure attention is paid to leave intact possible internal holes that are in nature the result of the superposition of leaf lobes or of different abiotic conditions.  
Note that this procedure must pay attention to leave internal holes intact, because in nature such holes are the result of the superposition of leaf lobes or due to several abiotic/biotic conditions.
Fig~\ref{fig:refin} shows the first leaf in Fig~\ref{fig:real_leaves} which presents artifacts (panel A, zoomed area including the artifact in panel B)
and its cleaned counterpart (panel C). 

\subsection*{Quantitative Quality Evaluation}

\begin{figure}[htbp]
\begin{center}
\includegraphics[width=\textwidth]{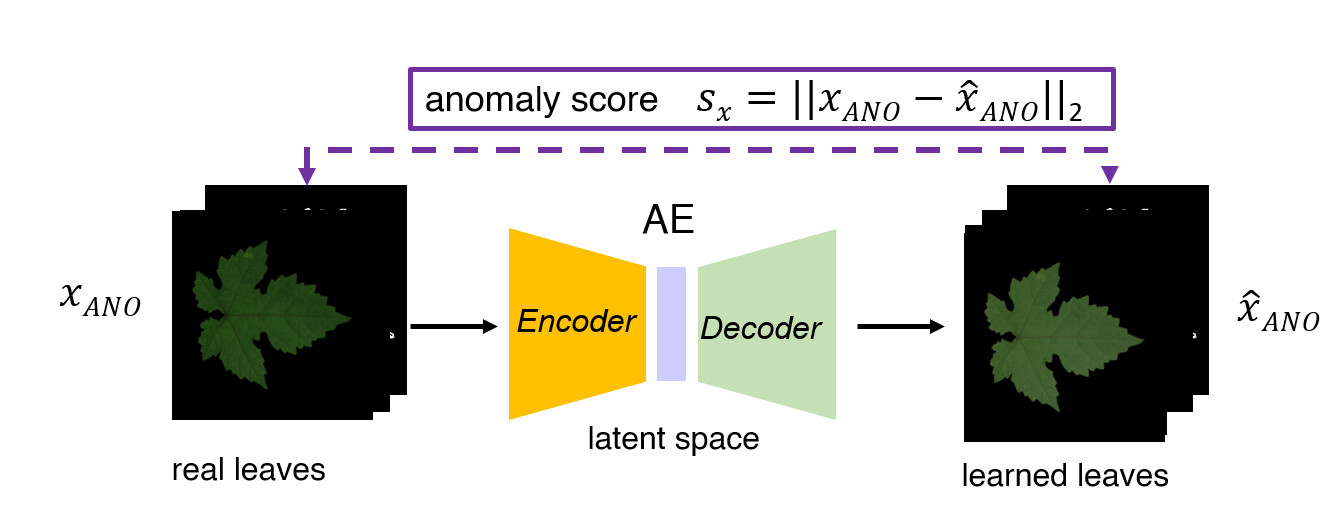}
\end{center}
\caption{{\bf AE for anomaly detection.} The AE is trained with images of real leaves
to be the identity operator of the input. A synthetic leaf with a low level of similarity is
recognized as an anomaly if fed into the trained AE and its anomaly score $s_x$ is high.}
\label{fig:AD}
\end{figure}

In order to assess quantitatively  the quality of the generated leaves, we employ a DL--based anomaly detection strategy.
This approach is discussed in detail in~\cite{benfenati2021unsupervised}, here we briefly recall the main points.
 The strategy consists in training an AE to compress {\em real} leaf images in a latent subspace and then reconstruct the images using the latent representation (see \textbf{Skeleton Generation section} for the same concept).  
Once the network is trained in this way, we feed it with a synthetic image
generated by our procedure. The AE encodes it in the latent space and tries to recover the original image according to its training rules. Since the net has been trained to be the identity operator for real images, if the artificial images are substantially different, an anomalous reconstruction
 is obtained.  
%An additional measure of the anomaly can be obtained as follows. After the first encoding/decoding is performed, another encoding is carried out. In this way, we can compare the latent variables obtained in the first step and the latent variables obtained in the second one: if they are similar by means of some suitable distance, then we can label the image as a real leaf image. Otherwise, if the latent variables are significantly different, then the image is labeled as synthetic. 
%A difference in latent variable representation stands for the fact that the first encoding/decoding process failed to recover the image, being trained to be the identity operator on real images. 
%Consequently, the second encoding will not extract the same latent features and an anomaly
%will be detected. 
Fig~\ref{fig:AD} provides a visual schematization of this approach.
The figure also details the score system used to detect the anomaly. 

%We trained an AE, named Model S3: we measured the relative difference $s_i$ between original image $x$ and decoded image $\hat{x}$ 
%$$
%s_i(x,\hat{x})= \frac{\|x-\hat{x}\|_2}{\|x\|_2}
%$$
%and the relative difference on the features of the two images, namely $f$ and $\hat{f}$:
%$$
%s_f(x,\hat{x}) =  \frac{\|f-\hat{f}\|_2}{\|f\|_2}
%$$
The degree of anomaly is quantified via the ROC curve and its area, the AUC index~\cite{bradley1997use}.
For this latter, we found AUC=0.25, which means that for a random synthetic image fed into the AE, 
there is a 25\% of possibility to classify it as an anomaly, that is to be synthetic instead or real. 
While this result does not indicate a perfect reconstruction of the real leaves, 
it shows that the synthetic leaves are a reasonably accurate surrogate of real
leaves and can be used for a first massive training at a very low cost. A successive refinement can then be
applied using a limited number of real leaves upon transfer learning. 

\begin{figure}
\begin{center}
\includegraphics[width=.5\textwidth]{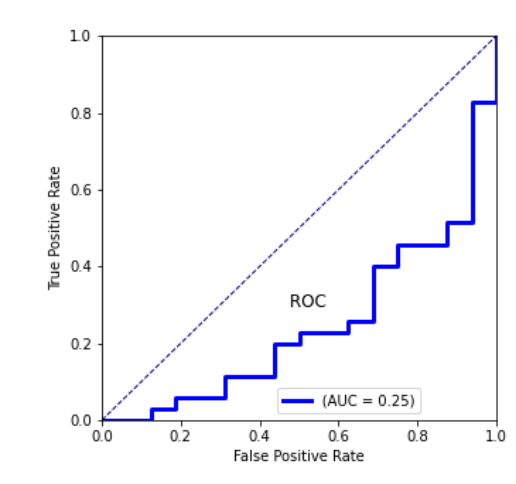}
\end{center}
\caption{{\bf Quantification of anomaly via ROC curve and AUC index.} A point on the ROC curve represents - for a certain threshold on the anomaly score  - the false positive rate (genuinely real images)
vs the true positive rate (genuinely synthetic images). The value AUC=0.25 means that
a synthetic image is (mis--)classified as synthetic in the 25\% of cases. The dotted line represents the result
one would obtain by tossing a coin to decide whether an image is artificial or real.}
\label{fig:ROC}
\end{figure}

\section*{Conclusion}

Goal of this work was to explore advanced DL generative methods to produce
realistic images of leaves to be used in computer--aided applications 
 The main focus was on the generation of artificial samples
 of leaves to be used to train DL networks for modern crop management systems
 in precision agriculture. 
 Disposing of synthetic samples which have a reasonable resemblance
 to real samples alleviates the burden of manually collecting and annotating 
 hundreds of data. 
The Pix2pix net performs good translations from the leaf skeletons
generated by the ResVAE, except for some discolored parts, both for the colorization
of RGB and RGNIR images. Also, the leaves
generated by ResVAE have sometimes pixels positioned outside the boundary
which, if not corrected, can cause artifacts in the synthetic leaves. 
An easy procedure has been proposed as well to correct these artifacts. 
We believe that the generative approach can significantly contribute to 
automatize the process of building a low-cost training set for
DL applications. Several computer--aided
applications may also benefit of such a strategy, where many samples are
required, possibly with different degree of accuracy in the representation.

\section*{Author contribution}

{\bf Conceptualization:} Alessandro Benfenati, Paola Causin \\[2mm]
\noindent {\bf Dataset:} Alessandro Benfenati, Davide Bolzi, Paola Causin, Roberto Oberti \\[2mm]
\noindent
{\bf Methodology:} Alessandro Benfenati, Davide Bolzi, Paola Causin \\[2mm]
\noindent
{\bf Implementation:}  Davide Bolzi\\[2mm]
\noindent
{\bf Analysis:} Alessandro Benfenati, Davide Bolzi, Paola Causin, Roberto Oberti \\[2mm]
\noindent
{\bf Writing:} Alessandro Benfenati, Paola Causin

\section*{Supporting information}
%
%% Include only the SI item label in the paragraph heading. Use the \nameref{label} command to cite SI items in the text.
\paragraph*{S1 Appendix}
\label{SI:ResVAE}
%{\bf Bold the title sentence.} Add descriptive text after the title of the item (optional).

\begin{figure}[ht]
\centering
\includegraphics[width=1\linewidth]{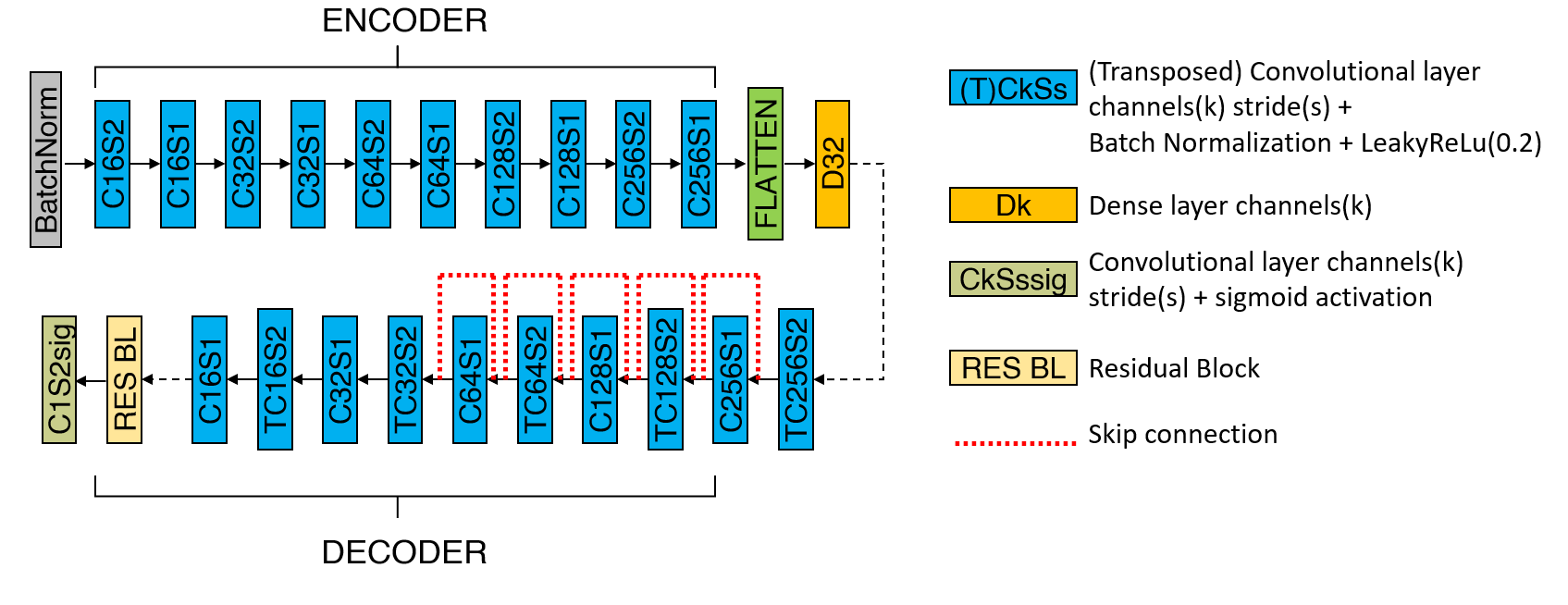}
\caption{{
\bf ResVAE architecture.} Building blocks of the encoder and decoder components of the ResVAE. The
convolutional filters have kernels of size $4\times4$. The Residual block is formed by 5 convolutional layers 
of 16 filters each with kernel of size $4\times4$ and stride equal to 1,
 followed by a Batch Normalization layer and LeakyReLU 
activation function.}
\label{fig:ResVAEarch}
\end{figure}

{\bf Implementation and training of the ResVAE neural network.} The architecture, inspired by the one described in~\cite{cholletResVAE},
is shown in Fig~\ref{fig:ResVAEarch}.
The training is performed via a stochastic gradient descent strategy,
with gradients computed by standard back--propagation; 
we use the Adam optimizer with  learning rate  $\eta=0.001$ and we train the model
for 2000 epochs with a batch size of 64. 
After a hyper--parameter search,
$\beta$ in the loss function~\eqref{eq:lossVAE} was set to 75.

\paragraph*{S2 Appendix}
{\bf Implementation and training of the {\tt Pix2pix} neural network.}
\label{SI:Pix2pix}

\begin{figure}[ht]
\centering
\includegraphics[width=1\linewidth]{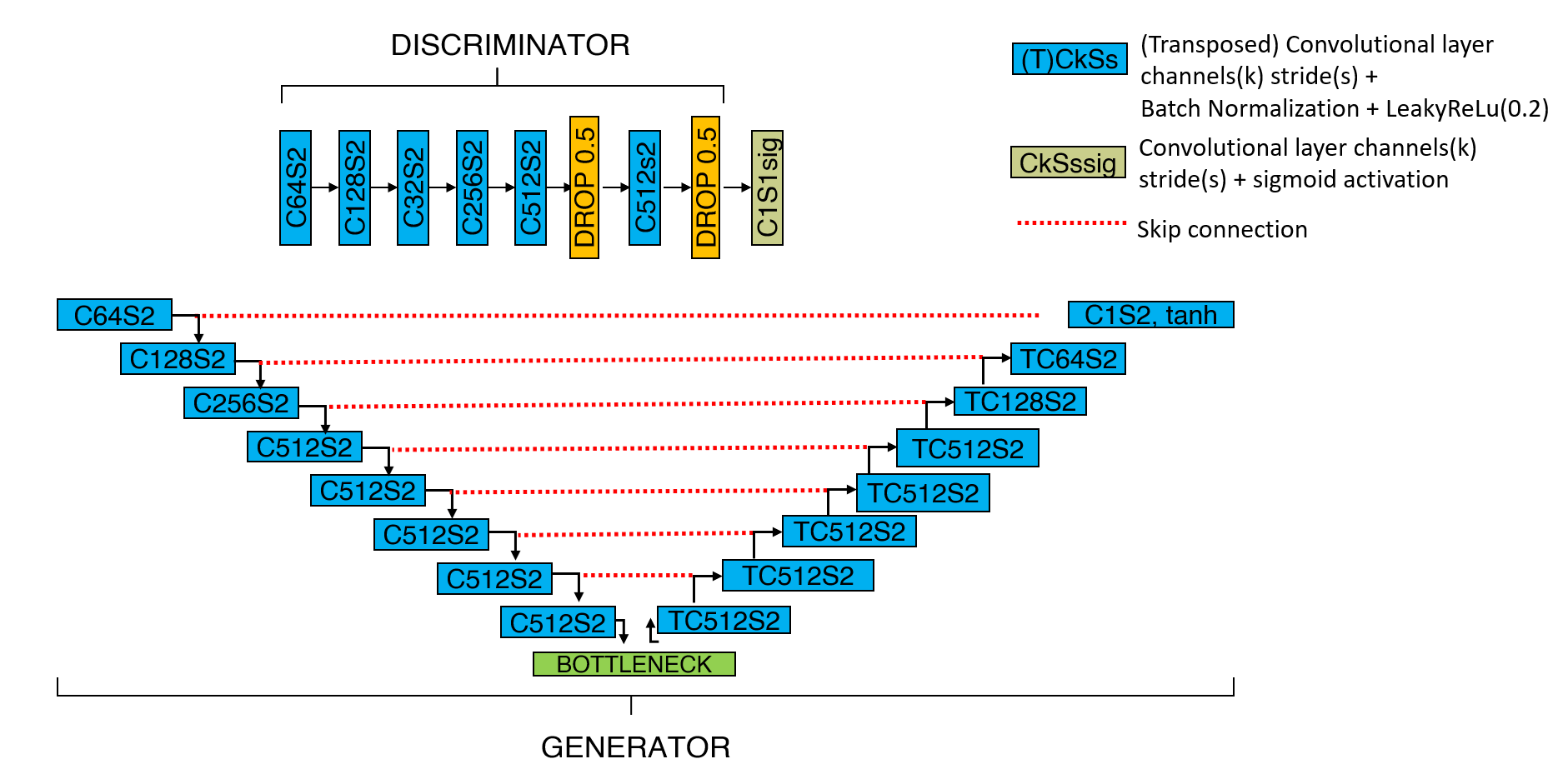}
\caption{
{\tt Pix2pix} {\bf architecture.} Building blocks of the generator and discriminator components.}
\label{fig:Pix2pixarch}
\end{figure}

The {\tt Pix2pix} is a GAN architecture designed for image-to-image translation,
originally presented in~\cite{isola2017image} and comprising a generator 
and a discriminator.  
The discriminator is deep neural network that performs image classification.  
It takes both the source image (leaf skeleton) and the target image (colorized leaf) as input and predicts the likelihood of whether the target image is real or a fake translation of the source image.
We use a PatchGAN model which tries to establish whether each $N \times N$  (local) patch in the image is real or fake. We run this discriminator convolutionally across the image, averaging all responses to provide the ultimate output of the discriminator.
The generator  is an encoder-decoder model using a U-Net architecture
with feature-map concatenation between two corresponding blocks of the encoder/decoder.
The encoder and decoder of the generator are comprised of standardized blocks of convolution, batch normalization, dropout, and activation layers.
We proceed as suggested in~\cite{isola2017image}: the generator is updated via a weighted sum of both the adversarial loss and the $L_1$ loss, where the parameter $\lambda$ in the loss function~eq\ref{eq:losscGANlambda} is set to 100 in order to encourage the generator to produce plausible translations of the input image, and not just plausible images 
in the target domain.
 We initialize the generator/discriminator weights with a normal distribution of zero mean and standard deviation $\sigma= 0.002$; we use the Adam optimizer with a learning rate $\eta=0.0002$ and 
we train the generator/discriminator paired model for 12000 training steps, using a batch size of~1.
Fig~\ref{fig:Pix2pixarch} shows the generator and discriminator architectures.

%\label{S1_Fig}
%{\bf Bold the title sentence.} Add descriptive text after the title of the item (optional).
%
%\paragraph*{S2 Fig.}
%\label{S2_Fig}
%{\bf Lorem ipsum.} Maecenas convallis mauris sit amet sem ultrices gravida. Etiam eget sapien nibh. Sed ac ipsum eget enim egestas ullamcorper nec euismod ligula. Curabitur fringilla pulvinar lectus consectetur pellentesque.
%
%\paragraph*{S1 File.}
%\label{S1_File}
%{\bf Lorem ipsum.}  Maecenas convallis mauris sit amet sem ultrices gravida. Etiam eget sapien nibh. Sed ac ipsum eget enim egestas ullamcorper nec euismod ligula. Curabitur fringilla pulvinar lectus consectetur pellentesque.
%
%\paragraph*{S1 Video.}
%\label{S1_Video}
%{\bf Lorem ipsum.}  Maecenas convallis mauris sit amet sem ultrices gravida. Etiam eget sapien nibh. Sed ac ipsum eget enim egestas ullamcorper nec euismod ligula. Curabitur fringilla pulvinar lectus consectetur pellentesque.
%
%\paragraph*{S1 Appendix.}
%\label{S1_Appendix}
%{\bf Lorem ipsum.} Maecenas convallis mauris sit amet sem ultrices gravida. Etiam eget sapien nibh. Sed ac ipsum eget enim egestas ullamcorper nec euismod ligula. Curabitur fringilla pulvinar lectus consectetur pellentesque.
%
%\paragraph*{S1 Table.}
%\label{S1_Table}
%{\bf Lorem ipsum.} Maecenas convallis mauris sit amet sem ultrices gravida. Etiam eget sapien nibh. Sed ac ipsum eget enim egestas ullamcorper nec euismod ligula. Curabitur fringilla pulvinar lectus consectetur pellentesque.

\section*{Acknowledgments}
We acknowledge support from the SEED PRECISION
project (PRecision crop protection: deep learnIng and data fuSION), funded by 
Universit\`a degli Studi di Milano. AB and PC are part of   
the GNCS group of INDAM (Istituto Nazionale di Alta Matematica "Francesco Severi").

%\nolinenumbers

% Either type in your references using

%
% or
%
% Compile your BiBTeX database using our plos2015.bst
% style file and paste the contents of your .bbl file
% here. See http://journals.plos.org/plosone/s/latex for 
% step-by-step instructions.
% 

%\bibliography{biblio}
%\bibliographystyle{vancouver}

\end{document}